\documentclass[lettersize,journal]{IEEEtran}
\usepackage{amsmath,amsfonts}
\usepackage{algorithmic}
\usepackage{algorithm}
\usepackage{array}
\usepackage{graphicx}

\usepackage{multirow}
\usepackage[capitalize]{cleveref}
\usepackage{booktabs}
\usepackage{pifont}
\usepackage{bm}
\usepackage[dvipsnames,table]{xcolor}
\definecolor{RowColor}{rgb}{0.95, 0.95, 1}

\newcommand{\cmark}{\textcolor{green}{\ding{51}}}%
\newcommand{\xmark}{\textcolor{red}{\ding{55}}}%

\newcommand{\RD}[1]{\textcolor{black}{#1}}

\begin{document}

\title{Efficient Hyperspectral Image Reconstruction Using Lightweight Separate Spectral Transformers}

\author{Jianan Li, Wangcai Zhao, Tingfa Xu$^{\dagger}$
        % <-this % stops a space
\thanks{$^{\dagger}$ Correspondence to: Tingfa Xu.}}% <-this % stops a space}

\markboth{Journal of \LaTeX\ Class Files,~Vol.~14, No.~8, August~2021}%
{Shell \MakeLowercase{\textit{et al.}}: A Sample Article Using IEEEtran.cls for IEEE Journals}


\maketitle

\begin{abstract}
Hyperspectral imaging (HSI) is essential across various disciplines for its capacity to capture rich spectral information. However, efficiently reconstructing hyperspectral images from compressive sensing measurements presents significant challenges. To tackle these, we adopt a divide-and-conquer strategy that capitalizes on the unique spectral and spatial characteristics of hyperspectral images. We introduce the Lightweight Separate Spectral Transformer (LSST), an innovative architecture tailored for efficient hyperspectral image reconstruction. This architecture consists of Separate Spectral Transformer Blocks (SSTB) for modeling spectral relationships and Lightweight Spatial Convolution Blocks (LSCB) for spatial processing. The SSTB employs Grouped Spectral Self-attention and a Spectrum Shuffle operation to effectively manage both local and non-local spectral relationships. Simultaneously, the LSCB utilizes depth-wise separable convolutions and strategic ordering to enhance spatial information processing. Furthermore, we implement the Focal Spectrum Loss, a novel loss weighting mechanism that dynamically adjusts during training to improve reconstruction across spectrally complex bands. Extensive testing demonstrates that our LSST achieves superior performance while requiring fewer FLOPs and parameters, underscoring its efficiency and effectiveness. The source code is available at: https://github.com/wcz1124/LSST.

\end{abstract}


\section{Introduction}
\IEEEPARstart{H}{yperspectral} imaging (HSI) is critical for a wide range of applications due to its ability to capture extensive spectral information~\cite{10342753,9034153,sun2019online}. The Coded aperture snapshot spectral imaging (CASSI) system employs a single-shot technique~\cite{gehm2007single} that encodes this information onto a two-dimensional sensor. However, reconstructing hyperspectral images from CASSI measurements presents considerable challenges, including the ill-posed nature of inverse problems and high computational demands~\cite{10604291,10261266,yao2024specat}.

Recent advancements in deep learning have significantly advanced the reconstruction of hyperspectral images. Deep Convolutional Neural Networks (CNNs) effectively convert 2D compressed and aliased images into 3D hyperspectral cubes. However, CNNs encounter challenges in spatial and spectral modeling, especially in capturing long-range dependencies. In contrast, Transformers~\cite{vaswani2017attention}, employing Multi-Head Self-Attention, excel at managing long-range dependencies and have emerged as a promising alternative~\cite{cai2022mask, cai2022coarse}.

\begin{figure}[t]
    \centering
    \includegraphics[width=0.5\textwidth]{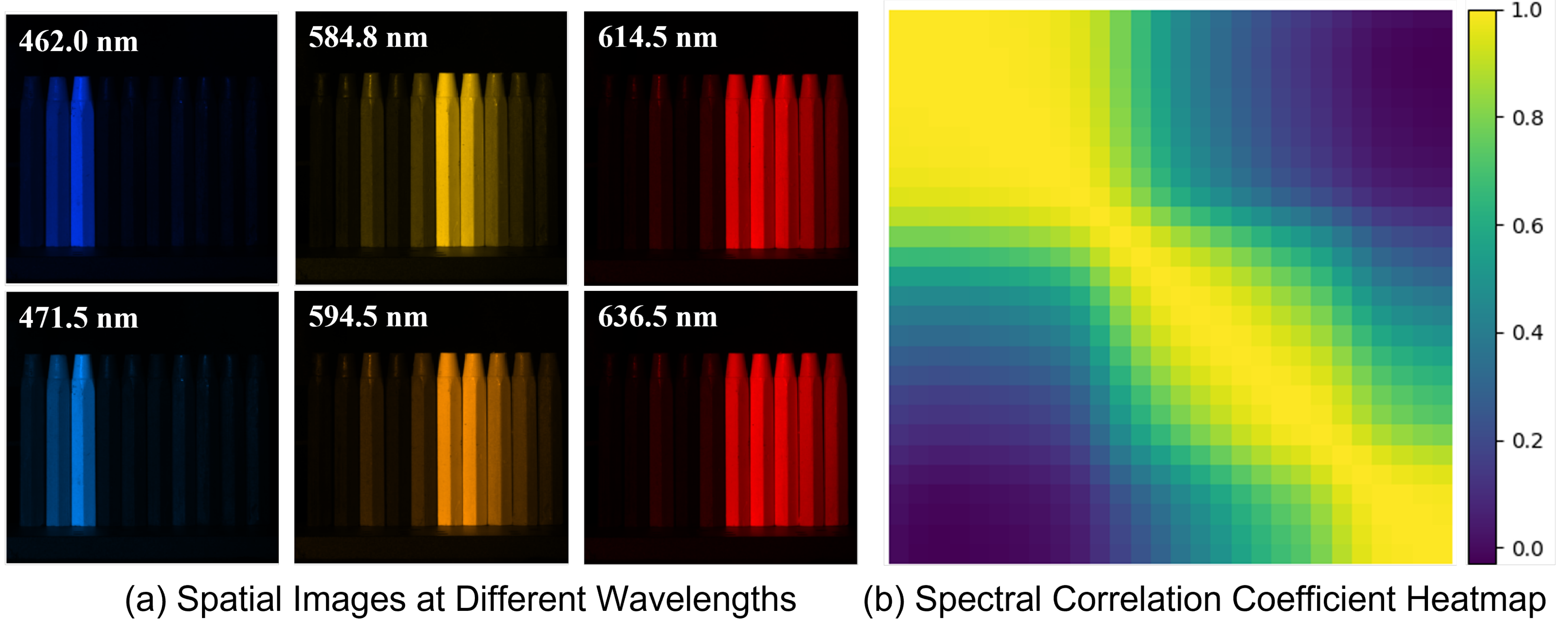}
    \vspace{-6mm}
    \caption{Natural hyperspectral images display unique properties. (a) In each spectral band, the local neighborhoods in the image typically exhibit stronger spatial correlation. (b) Spectral bands that are closer in proximity often demonstrate stronger correlations compared to those that are more distant.}
    \label{fig:Motivation}
\end{figure}

\RD{We observe that natural hyperspectral images exhibit distinctive statistical characteristics. First, local neighborhoods within an image typically correspond to similar materials, resulting in highly consistent spectral signatures. As illustrated in \cref{fig:Motivation}(a), such regions present much stronger spatial correlations within each spectral band, suggesting that global spatial modeling provides limited additional benefit. Second, \cref{fig:Motivation}(b) shows that spectral correlations are concentrated near the diagonal, while correlations decay rapidly toward the off-diagonal regions. This indicates that adjacent spectral bands are far more strongly correlated than distant ones.}

\RD{Given these properties, directly applying the original Transformer, which uniformly emphasizes global spatial and spectral dependencies, may not be optimal for hyperspectral image reconstruction. Moreover, the quadratic computational complexity of global attention with respect to both spectral and spatial dimensions introduces substantial computational overhead, posing practical challenges for lightweight or resource-constrained applications.}

To address these challenges, we introduce the Lightweight Separate Spectral Transformer (LSST), a highly efficient Transformer architecture tailored for hyperspectral image reconstruction. The LSST features a U-shaped configuration composed of consecutive Lightweight Separate Spectral Transformer Blocks. Each block houses two key components: the Separate Spectral Transformer Block (SSTB) and the Lightweight Spatial Convolution Block (LSCB). These components employ a divide-and-conquer approach to distinctly model the spectral and spatial relationships, efficiently capitalizing on the unique characteristics of hyperspectral images across both dimensions.

In modeling spectral relationships, it's crucial to recognize that bands closer together often exhibit stronger correlations than distant ones. Applying global attention indiscriminately across all spectral bands poses two challenges: it may overlook critical local spectral dynamics and impose substantial computational burdens due to the extensive range of global attention.

To overcome these issues, the Separate Spectral Transformer Block employs a strategic phased approach that captures both local and non-local spectral dependencies with improved computational efficiency. Specifically, it first segments the input feature map into clusters along the spectral dimension. Grouped Spectral Self-attention is then applied within each cluster to target local correlations. To efficiently model non-local relationships, the spectral bands of the resulting feature map undergo a parameter-free Spectrum Shuffle operation, which facilitates the intermixing of distant bands. This setup allows subsequent Grouped Spectral Self-attention to effectively capture non-local spectral relationships across the entire spectrum. Consequently, the Separate Spectral Transformer Block effectively captures comprehensive spectral interactions solely through efficient local attention operations.

\begin{figure}[t]
    \centering
    \includegraphics[width=0.40\textwidth]{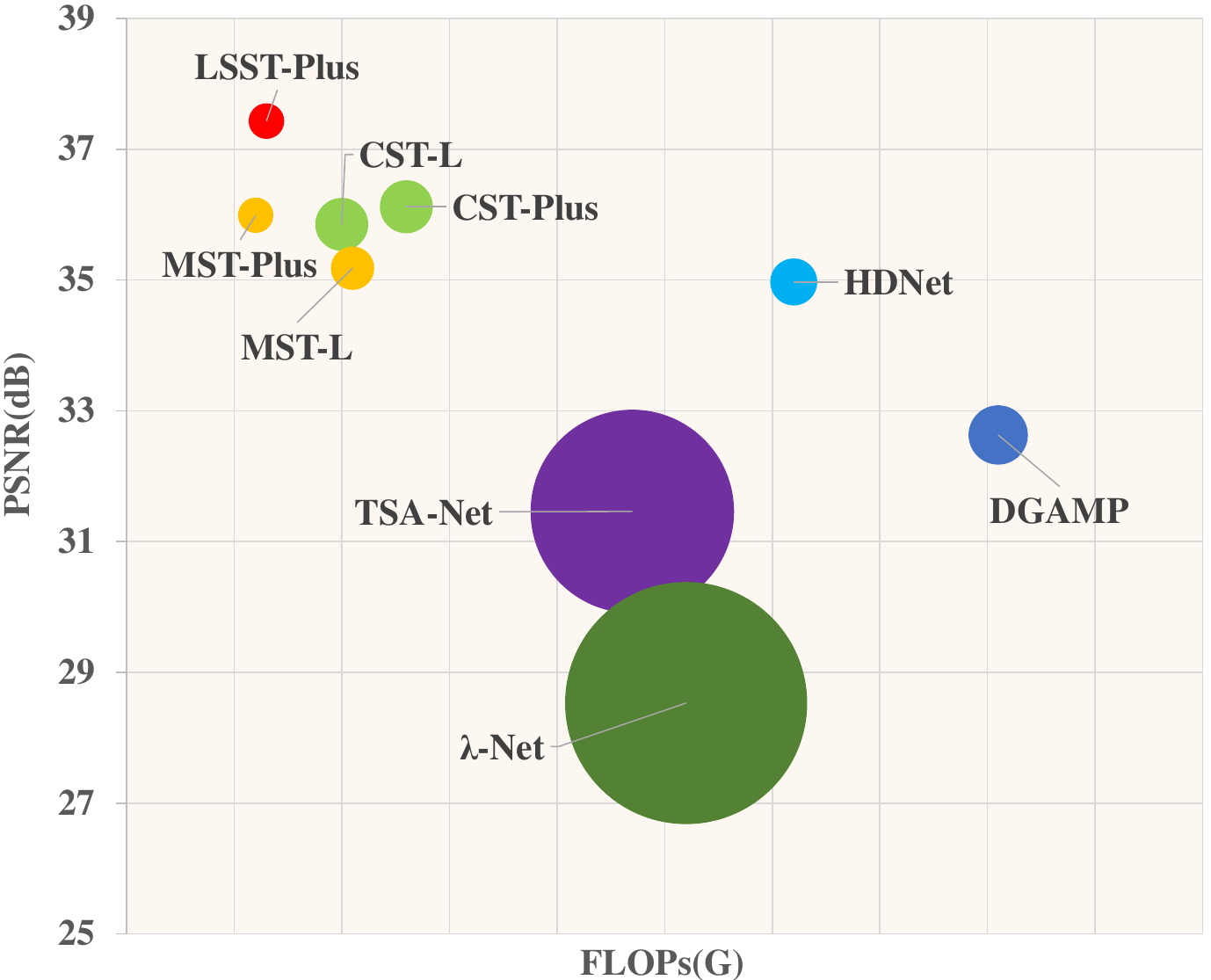}
    \vspace{-2mm}
    \caption{Comparing accuracy and efficiency among various approaches. The circle's radius represents the number of model parameters.}
    \vspace{-2mm}
    \label{fig:Comparison}
\end{figure}

In modeling spatial relationships, inspired by ConvNeX~\cite{liu2022convnet}, the Lightweight Spatial Convolution Block (LSCB) replaces global attention in the spatial dimension with depth-wise separable convolution using a large kernel size, efficiently capturing local spatial relationships. Diverging from conventional configurations of convolutional modules, the LSCB strategically positions depth-wise separable convolutions before channel-expanding convolutions, substantially reducing the number of parameters and decreasing the computational load.

Moreover, variability among spectral bands results in distinct challenges for image reconstruction, as treating all bands uniformly can bias the model towards those that are simpler to reconstruct, thus neglecting the more complex bands. To address this imbalance, we take inspiration from Lin et al.~\cite{Lin_2017_ICCV} and implement a novel approach called Focal Spectrum Loss. This technique assesses reconstruction quality across different bands during training and adjusts the loss weights dynamically. This adaptive mechanism ensures that the model attentively enhances reconstruction across all spectral bands, thereby improving overall image quality.

Extensive experiments underscore the superior performance of our LSST method. As depicted in~\cref{fig:Comparison}, LSST outperforms existing methods while requiring fewer FLOPs and parameters. This demonstrates the exceptional efficiency and effectiveness of our approach.

To summarize, this work makes the following contributions:
\begin{itemize}
\item The Lightweight Separate Spectral Transformer utilizes a divide-and-conquer strategy to model spectral and spatial relationships efficiently, reducing computational needs while maintaining high reconstruction quality.
\item The SSTB employs a novel phased approach with Grouped Spectral Self-attention and Spectrum Shuffle, effectively managing local and non-local spectral dependencies and minimizing computational overhead.
\item The LSCB utilizes depth-wise separable convolution and strategic ordering to reduce computational complexity while enhancing spatial analysis.
\item The introduction of Focal Spectrum Loss dynamically adjusts loss weights based on spectral reconstruction quality during training, promoting balanced enhancement across all spectral bands and improving overall image quality.
\end{itemize}

\section{Related Work}

\subsection{HSI Reconstruction.}
Recent advancements in HSI reconstruction from compressed measurements through CASSI systems have marked significant progress. Initially, HSI reconstruction relied on model-based methods that utilized hand-crafted image priors to constrain solutions within desired data spaces~\cite{yuan2016generalized, liu2018rank}. These methods, however, required manual parameter adjustments, which not only slowed the reconstruction process but also limited their representational capacity and generalizability. More recently, Convolutional Neural Networks (CNNs) have been employed to solve the inverse problem of spectral snapshot compressive imaging. These CNN-based approaches are categorized into three main types: end-to-end (E2E) methods~\cite{miao2019net, meng2020end}, deep unfolding methods~\cite{wang2019hyperspectral, wang2020dnu, huang2021deep, ma2019deep}, and plug-and-play (PnP) methods~\cite{zheng2021deep, yuan2020plug}.

In light of these developments, creating lightweight models for HSI reconstruction has become crucial for real-time applications. Recent efforts~\cite{cai2022coarse, li2023pixel} have focused on designing such models to enhance reconstruction efficiency without compromising accuracy. Building on these innovations, this work introduces specialized design elements in both spatial and spectral dimensions, aimed at reducing model complexity and computational demands, thus making it particularly well-suited for deployment in resource-constrained environments.

\subsection{Transformers for HSI Reconstruction}
In recent years, the application of Transformer models~\cite{vaswani2017attention} to enhance the reconstruction quality of HSIs has attracted considerable attention. TSA-Net~\cite{meng2020end} introduced spatial-spectral self-attention mechanisms to sequentially reconstruct HSIs. Significantly, MST~\cite{cai2022mask} conceptualizes spectral bands as tokens and applies self-attention across the spectral dimension. Further exploring the capabilities of Transformers in compressive sensing, CST~\cite{cai2022coarse} was developed to exploit the inherent similarities within HSIs. Lin et al.~\cite{lin2023metasurface} augment the U-Net architecture with a novel Transformer block that reorders depthwise and standard convolutions, thereby improving feature-map interactions and attention efficiency while reducing the model’s parameter count. Additionally, PADUT~\cite{li2023pixel} introduced the Non-local Spectral Transformer, emphasizing the three-dimensional attributes of HSIs for enhanced recovery. \RD{In contrast to these approaches, which primarily emphasize modeling long-range spectral dependencies, our method is motivated by the observation that hyperspectral images exhibit strong local but comparatively weaker global correlations across spectral bands. Accordingly, we introduce the Separate Spectral Multi-head Self-Attention (SS-MSA), which explicitly captures both localized and non-local spectral interactions, providing a more targeted and effective mechanism for hyperspectral image reconstruction.}

\subsection{Loss Functions for HSI Reconstruction}
In the domain of HSI reconstruction, traditional methodologies typically treat the entire image, encompassing both spatial and spectral dimensions, as a unified entity. These methods predominantly utilize Root Mean Square Error (RMSE) as the principal criterion for training. However, this approach may not fully capture the complexities of HSI data. Recent advancements have seen a shift towards devising loss functions that are more attuned to the unique properties of HSIs. For instance, Song et al.~\cite{cai2022coarse} developed a learning-based algorithm that includes a sparsity loss to exploit the inherent spatial sparsity of HSIs. Hu et al.~\cite{9879297} implemented a loss function based on frequency domain analysis, offering an alternative method for processing HSIs by concentrating on their frequency attributes. Despite these innovations, challenges persist in effectively reconstructing the varied complexities of different spectral bands in HSIs, often leading to sub-optimal outcomes in the more complex bands. To address this, our study introduces the Focal Spectrum Loss, which dynamically adjusts the weights assigned to different bands during the training process. This method ensures a comprehensive and efficacious reconstruction across all spectral bands, markedly improving the quality and applicability of the reconstructed images.

\section{Method}

\subsection{Problem Formulation}
The Compressive Sensing Spectral Imaging (CASSI) system leverages the principles of compressive sensing to efficiently capture hyperspectral images. The operation of CASSI involves encoding spatial and spectral information onto a two-dimensional detector, utilizing a coded aperture and a dispersive element. The CASSI system captures encoded measurements $\bm y$ which can be modeled by the equation:
\begin{equation}
  \bm y = \bm \Phi \bm x + \epsilon.
\end{equation}
Here, $\bm x$ represents the original hyperspectral data, $\bm \Phi$ denotes the measurement matrix derived from the coded aperture and dispersive element, and $\epsilon$ is the noise. The primary task in HSI reconstruction is to recover $\bm x$ given $\bm y$ and $\bm \Phi$.

Reconstruction is facilitated by algorithms that capitalize on the sparsity of hyperspectral data in the spectral domain. Typically, the reconstruction challenge is expressed as:
\begin{equation}
  \hat{\bm x} = \arg \min_{\bm x} \|\bm \Phi \bm x - \bm y\|_2^2 + \lambda \|\bm x\|_1.
\end{equation}
In this formulation, $\lambda$ is a regularization parameter that helps balance between the fidelity to the measured data and the sparsity of the solution.

\begin{figure*}[t]
  \centering
    \includegraphics[width=0.75\textwidth]{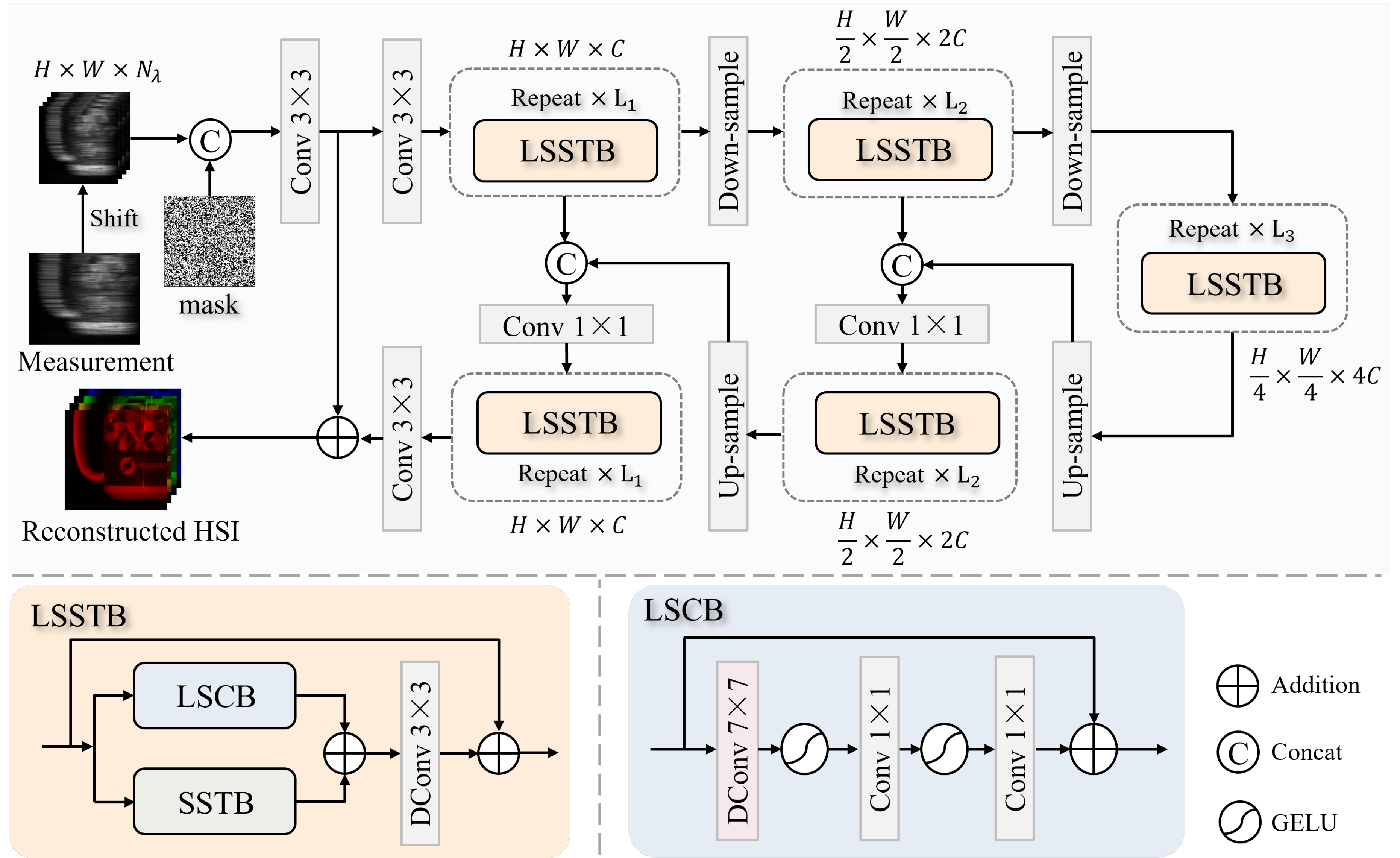}
    \vspace{-1mm}
    \caption{The overarching architecture of the LSST utilizes a U-shaped configuration, incorporating stacked Lightweight Separate Spectral Transformer Blocks (LSSTB). Each LSSTB comprises two primary components: the Separate Spectral Transformer Block (SSTB) and the Lightweight Spatial Convolution Block (LSCB), which are dedicated to efficiently modeling spectral and spatial relationships, respectively.}
   \label{fig:Overall}
\end{figure*}

\subsection{Overall Architecture}

\cref{fig:Overall} illustrates the LSST architecture, which adopts a U-shaped configuration consisting of encoder, bottleneck, and decoder stages. The system utilizes Lightweight Separate Spectral Transformer Blocks (LSSTB) arranged in a hierarchical structure. Initially, LSST reverses the dispersion process, converting the two-dimensional mixed-snapshot image $\bm y$ into an initialized spectral image $\bm X_0 \in \mathbb{R}^{{\rm H} \times {\rm W} \times {\rm N}_{\lambda}}$. This image has a spatial resolution of ${\rm H} \times {\rm W}$ and contains ${\rm N}_{\lambda}$ spectral bands. The image is then concatenated with the mask from the CASSI system and processed through $3 \times 3$ convolutions to extract shallow features, denoted by $\bm X \in \mathbb{R}^{{\rm H} \times {\rm W} \times {\rm C}}$, where ${\rm C}$ represents the channel dimension.

The encoder stage processes $\bm X$, incorporating two LSSTB modules, each followed by a down-sampling operation. Each down-sampling step reduces the spatial resolution by half and doubles the channel count. The bottleneck layer follows, consisting of a single LSSTB module. At this point, the input features retain their dimension and spatial resolution. Subsequently, the decoder stage methodically reconstructs the detailed structure of the hyperspectral image through two phases of upsampling and LSSTB modules, incrementally restoring the original size and details of the input image.

Mirroring the U-Net architecture, LSST integrates skip connections that link feature maps from the encoder and decoder stages. These connections enable the transfer of low-level feature information, significantly enhancing the network's ability to reconstruct detailed images.

LSSTB constitutes the core module within the LSST architecture, as depicted in~\cref{fig:Overall}. Each LSSTB is composed of two principal components: the Separate Spectral Transformer Block (SSTB) and the Lightweight Spatial Convolution Block (LSCB). The input features to LSSTB are concurrently transformed by the SSTB and LSCB modules, which focus on spectral and spatial relationship modeling, respectively. The outputs from both blocks are combined and subsequently processed through a $3 \times 3$ depth-wise separable convolution. \RD{While the combination can be implemented by channel-wise concatenation with subsequent projection, or by element-wise summation, we adopt the latter to preserve architectural simplicity.} This is followed by the addition of the original input to produce the final output. We next elaborate further on the SSTB and LSCB, respectively.

\subsection{Separate Spectral Transformer Block}
\begin{figure*}[t]
  \centering
    \includegraphics[width=0.9\textwidth]{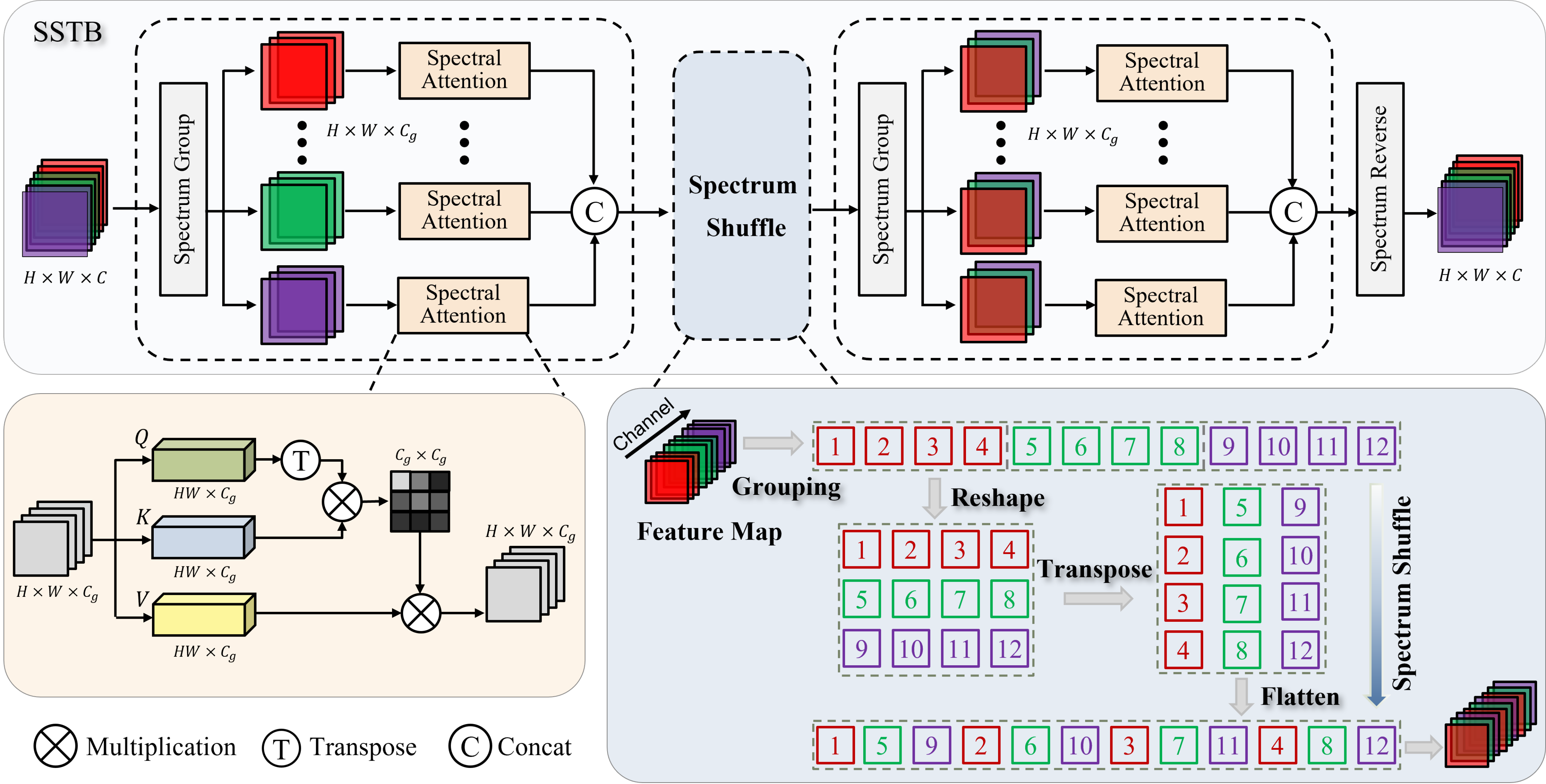}
    \caption{The conceptual framework of the Separate Spectral Transformer Block (SSTB) utilizes Grouped Spectral Self-attention along with a parameter-free Spectrum Shuffle operation, effectively managing both local and non-local spectral relationships.}
   \label{fig:Chart_SSTB}
\end{figure*}

\RD{\cref{fig:Chart_SSTB} illustrates the conceptual framework of the Separate Spectral Transformer Block (SSTB), which focuses attention computations within the spectral dimension. However, directly applying global attention to all spectral bands raises two key challenges. First, adjacent spectral bands typically exhibit stronger correlations than distant ones, and uniform modeling may overlook critical local spectral dynamics. Second, global attention over the entire spectral range incurs high computational cost. To address these issues, we adopt a phased strategy termed the Separate Spectral Multi-head Self-Attention mechanism (SS-MSA), which combines Local Spectral Self-Attention and Non-Local Spectral Self-Attention to capture both local and long-range spectral dependencies while improving computational efficiency.}

\subsubsection{Local Spectral Attention}
\RD{Given the input feature map $\bm X \in \mathbb{R}^{{\rm H} \times {\rm W} \times {\rm C}}$, this step aims to model local spectral correlations. Instead of applying attention across all wavelength channels, the feature map is first segmented along the spectral dimension into clusters via spectrum grouping. Grouped Spectral Self-Attention is then performed within each cluster to effectively capture local spectral correlations.}

\noindent\textbf{Spectrum Grouping.}
\RD{The input feature map is uniformly divided into groups along the spectral dimension, each containing ${\rm C}_g$ channels. The grouped features are defined as:
\begin{equation}
    \bm X_G = \{ \bm x_g \in \mathbb{R}^{{\rm H}{\rm W} \times {\rm C}_g} \ | \  g=1, \cdots, {\rm G} \},
\end{equation}
where ${\rm G}$ is the number of groups and ${\rm C}_g = {\rm C} / {\rm G}$ denotes the number of spectral bands per group.}

\noindent\textbf{Grouped Spectral Self-Attention.}
Following spectrum grouping, Spectral Self-Attention is applied within each group to model local spectral correlations. For the $g$-th group, $\bm x_g$ is transformed into the Query, Key, and Value vectors $\bm Q_g$, $\bm K_g$, and $\bm V_g$:
\begin{equation}
    \bm Q_g = \bm x_g \bm W^{Q}, \ \bm K_{g} = \bm x_g \bm W^{K}, \ \bm V_{g} = \bm x_g \bm W^{V},
\end{equation}
where $\bm W^{Q}$, $\bm W^{K}$, and $\bm W^{V}$ $\in \mathbb{R}^{{\rm C}_g \times {\rm C}_g}$ are learnable weight matrix. The self-attention operation updates the features as follows:
\begin{equation}
    \bm x_g' = \mathrm{Softmax}\left(\frac{\bm Q_g \bm K_g^{\top}}{\sqrt{{\rm C}_g}}\right) \bm V_g,
\end{equation}
The enhanced features from all groups are subsequently concatenated along the spectral dimension to reconstruct the updated feature map:
\begin{equation}
    \bm X' = \mathrm{Concat}\left\{\bm x_g'  \ | \  g=1, \cdots, {\rm G} \right\} \in \mathbb{R}^{{\rm H} {\rm W} \times {\rm C}}.
\end{equation}
This method effectively delineates local spectral correlations through grouped attention.

\subsubsection{Non-local Spectral Attention}
\RD{This phase aims to capture long-range dependencies across spectral dimensions. Inspired by ShuffleNet~\cite{zhang2018shufflenet}, we introduce an efficient Non-local Spectral Attention mechanism that combines Local Spectral Attention with Spectral Shuffle and Spectrum Reverse operations to effectively model non-local spectral dependencies.}

\noindent\textbf{Spectrum Shuffle.}
\RD{To efficiently model non-local spectral relationships using local spectral attention, we first rearrange the spectral bands of the input feature map $\bm X' \in \mathbb{R}^{{\rm H}{\rm W} \times {\rm C}}$ through a sequence of simple matrix operations, as shown in \cref{fig:Chart_SSTB}.}

\RD{The input feature map is first uniformly divided into ${\rm G}$ groups along the spectral dimension, each containing ${\rm C}_g$ channels, yielding an intermediate tensor of size ${\rm H}{\rm W} \times {\rm G} \times {\rm C}_g$. This tensor is then transposed to ${\rm H}{\rm W} \times {\rm C}_g \times {\rm G}$ and flattened to generate the spectrally shuffled feature map $\bm X'_S \in \mathbb{R}^{{\rm H}{\rm W} \times {\rm C}}$. This procedure enables effective non-local mixing of spectral bands.}

\noindent\textbf{Local Spectral Attention.}
\RD{Following Spectrum Shuffle, the feature map $\bm X'S$ is reorganized along the spectral dimension. Based on this configuration, we apply Local Spectral Attention (LSA) as described earlier, including Spectrum Grouping and Grouped Spectral Self-Attention:
\begin{equation}
    \bm X_{spe} = LSA(\bm X'_S).
\end{equation}
Spectrum Shuffle mixes spectral bands such that, after Spectrum Grouping, each group contains spectra spanning a wide wavelength range. Thus, self-attention within each group effectively captures non-local spectral dependencies across the entire spectrum.}

\noindent\textbf{Spectrum Reverse.}
\RD{To conclude the process, we apply the Spectrum Reverse operation, the inverse of Spectrum Shuffle, to $\bm X''$. This step restores the original spectral order of the feature map, yielding features that effectively capture both local and non-local spectral relationships.}

\subsubsection{Computational Overhead Analysis}
\RD{To demonstrate the lightweight design of our approach, we analyze the computational complexity of the proposed Separate Spectral Multi-head Self-Attention (SS-MSA) mechanism and compare it with several other Multi-head Self-Attention (MSA) variants. Consider an input feature map with spatial resolution $\rm H \times \rm W$ and $\rm C$ channels. The computational complexities are summarized as follows:}

\RD{G-MSA, the original global MSA defined in~\cite{dosovitskiy2020image}, scales quadratically with respect to ${\rm H} {\rm W}$:}
\begin{equation}
    O(\text{G-MSA}) = 2({\rm H}{\rm W})^2{\rm C}.
\end{equation}

\RD{The window-based self-attention (W-MSA)~\cite{liu2021swin} and spectral self-attention (S-MSA) are given by:
\begin{equation}
    O(\text{W-MSA}) = 2{\rm M}^2{\rm H}{\rm W} \cdot {\rm C} = 2{\rm M}^2{\rm H}{\rm W}{\rm C}.
\end{equation}
\begin{equation}
    O(\text{S-MSA}) = 2{\rm H}{\rm W}{\rm C}^2.
\end{equation}
Here, ${\rm M}$ denotes the local window size. Both W-MSA and S-MSA require substantially fewer computations compared to G-MSA, scaling linearly with the spatial size ${\rm H}{\rm W}$. However, the complexity of S-MSA grows quadratically with respect to the spectral dimension ${\rm C}$, resulting in significant computational overhead.}

\RD{In contrast, the computational complexity of our separate spectral attention is:
\begin{equation}
    O(\text{SS-MSA}) = 2{\rm H}{\rm W} ({\rm C}_g)^2 \cdot \frac{{\rm C}}{{\rm C}_g} = 2{\rm H}{\rm W}{\rm C}_g{\rm C}.
\end{equation}
Thus, SS-MSA achieves linear computational complexity with respect to both the spatial dimensions ${\rm H}{\rm W}$ and spectral dimension ${\rm C}$, significantly reducing computational cost. This highlights the efficiency of the proposed method in managing computational resources.}

\subsection{Lightweight Spatial Convolution Block}
\RD{To model spatial relationships within the feature map, we incorporate a Lightweight Spatial Convolution Block (LSCB), as shown in~\cref{fig:Overall}. The processing of the input feature map $\bm X \in \mathbb{R}^{{\rm H} \times {\rm W} \times {\rm C}}$ is formulated as:
\begin{equation}
    \bm X_{spa} = \bm X + \operatorname{Conv}_2\left(\bm \sigma\left(\operatorname{Conv}_1(\bm \sigma(\operatorname{DWConv}(\bm X)))\right)\right).
\end{equation}
Here, $\operatorname{DWConv}$ denotes depth-wise convolution with a $7\times7$ kernel for capturing local spatial details. $\operatorname{Conv}_1$ is a $1\times1$ convolution that linearly projects the feature from ${\rm C}$ to $4{\rm C}$ channels, and $\operatorname{Conv}_2$ reduces it back to ${\rm C}$. The function $\bm \sigma$ represents the GELU activation.}

\RD{This block rearranges the conventional convolutional structure by placing depth-wise separable convolution before the $1\times 1$ expansion. This design reduces both parameters and computation, enabling LSCB to efficiently model spatial relationships.}

\subsection{Focal Spectrum Loss}
\begin{figure}[t]
    \centering
    \includegraphics[width=0.5\textwidth]{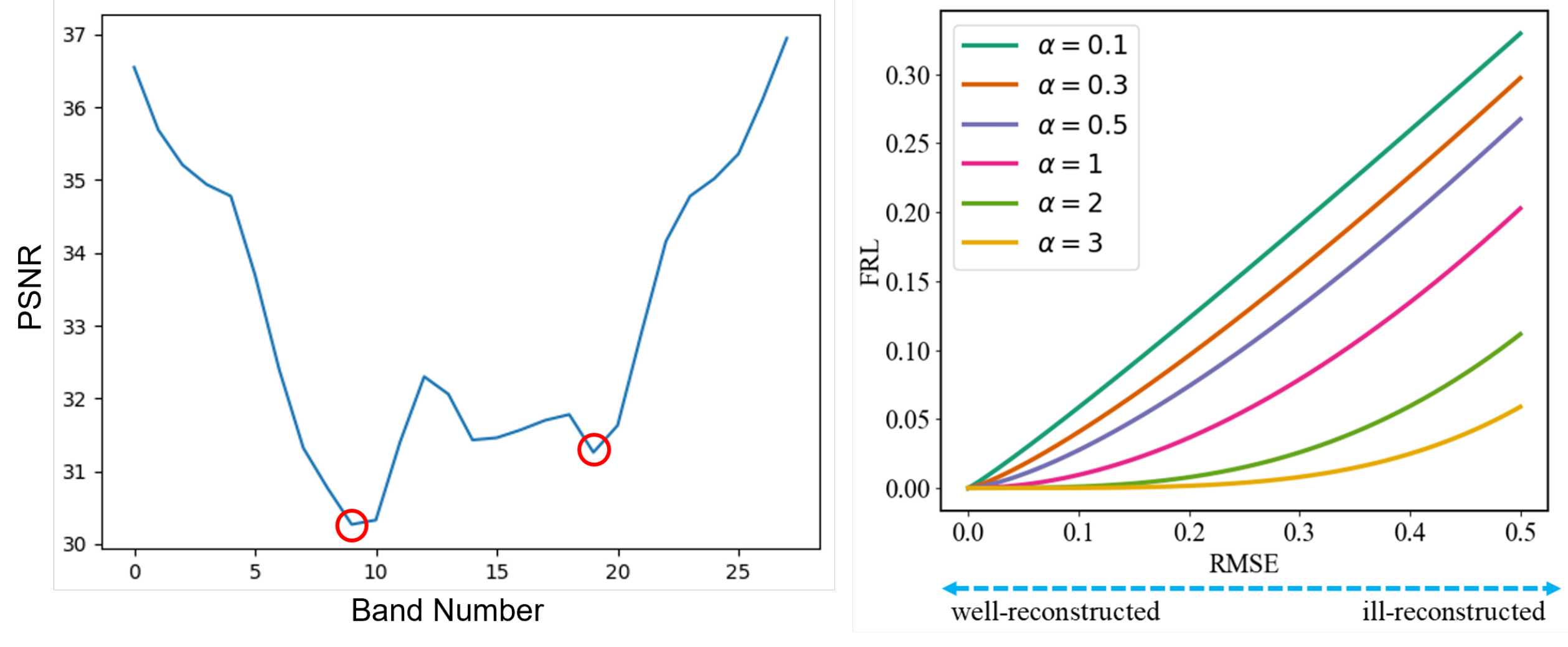}
    \vspace{-6mm}
    \caption{Left: PSNR for reconstructed results across various frequency bands. Right: Focal Spectrum Loss with different focusing parameters.}
    \vspace{-2mm}
    \label{fig:Vis_Loss}
\end{figure}

\RD{Conventional reconstruction methods primarily employ Root Mean Square Error (RMSE) as their optimization objective. This approach minimizes the global RMSE between the reconstructed hyperspectral image and its ground truth, thereby improving the overall reconstruction quality. However, hyperspectral images exhibit distinct spectral signatures across different bands, resulting in varying levels of reconstruction difficulty. As illustrated on the left in~\cref{fig:Vis_Loss}, certain bands—such as those highlighted by the red circle—are intrinsically more challenging to reconstruct under spectral compression. When a single global RMSE is optimized uniformly across all spectral bands, the model tends to prioritize learning bands that are easier to reconstruct, while neglecting complex bands that contain richer spectral information. This imbalance ultimately leads to degraded fidelity in spectral reconstruction, especially in the most informative yet difficult bands.}

\RD{To mitigate this issue, and inspired by Focal Loss~\cite{Lin_2017_ICCV}, we propose a novel loss function termed Focal Spectrum Loss (FSL). During training, FSL computes the RMSE for each spectral band individually, where higher RMSE values naturally correspond to bands with greater reconstruction complexity. Based on this observation, FSL adaptively assigns larger loss weights to these “hard-to-reconstruct” bands, guiding the model to allocate more learning capacity to bands that have not yet been sufficiently recovered. Rather than treating all bands equally, FSL continuously monitors the per-band reconstruction quality and dynamically updates the loss weights over the course of optimization.}

\RD{Intuitively, the FSL mechanism can be viewed as a spectrum-aware focusing strategy: the per-band RMSE serves as a measure of reconstruction difficulty, and the loss amplifies the contribution of bands with larger errors. This allows the model to actively correct residual spectral distortion where it matters most, rather than being dominated by easier patterns. As a result, the model adaptively emphasizes complex spectral regions and achieves a more balanced and complete reconstruction of all spectral bands.}

Let $\bm{Y}$ and $\bm{\hat{Y}}$ denote the reconstructed and ground-truth hyperspectral images, respectively, both situated within the space $\mathbb{R}^{{\rm H}\times {\rm W} \times {\rm N_{\lambda}}}$. Here, ${\rm H} \times {\rm W}$ represents the spatial dimensions, and ${\rm N_{\lambda}}$ signifies the number of spectral bands. The index $k$ ranges from 1 to ${\rm N_{\lambda}}$, each corresponding to a distinct spectral band, with $\bm{Y}_k$ and $\bm{\hat{Y}}_k$ in $\mathbb{R}^{{\rm H} \times {\rm W}}$ representing the reconstructed and ground-truth HSI for the $k$-th band, respectively. The Focal Spectrum Loss (FSL) function is formulated as:
\begin{equation}
    \mathcal{L}_{FSL} = \frac{1}{{\rm N}_{\lambda}} \sum_{k=1}^{{\rm N}_{\lambda }} \omega_k \bm \ell_k,
\end{equation}
where $\bm \ell_{k}$ denotes the RMSE loss for the $k$-th spectral band, computed as:
\begin{equation}
    \bm \ell_k=\sqrt{\frac{1}{{\rm H}{\rm W}} \sum_{i=1}^{{\rm H}{\rm W}}\left(\bm{Y}_{k,i}-\bm{\hat{Y}}_{k,i}\right)^{2}},
\end{equation}
with $\bm{Y}_{k,i}$ and $\bm{\hat{Y}}_{k,i}$ representing the $i$-th pixel in the $k$-th band of the reconstructed and ground-truth images, respectively. The dynamic weight $\omega_{k}$ for the $k$-th band is given by:
\begin{equation}
\omega_{k}=\log((\bm \ell_{k})^{\alpha}+1),
\end{equation}
where $\alpha > 0$ is the focal parameter enhancing the focus on more challenging spectral bands.

\cref{fig:Vis_Loss} on the right demonstrates the effects of Focal Spectrum Loss within the range [0.1, 3] for varying values of $\alpha$. The adjustment factor $\omega_{k}$ is pivotal in modulating the relative losses across different spectral bands. Specifically, for well-reconstructed spectra where $\bm \ell_{k}$ is low, $\omega_{k}$ approaches zero, thereby minimizing the loss weight for these spectra. Conversely, as $\bm \ell_{k}$ increases, indicating poorer reconstruction, $\omega_{k}$ also increases, thereby accentuating the loss for more challenging spectra. The focal parameter $\alpha$ finely tunes the rate at which these adjustments occur, with higher $\alpha$ values intensifying the influence of $\omega_{k}$. Experimental results presented in this study suggest that an $\alpha$ value of 0.5 delivers optimal results.

\begin{table*}[t]
\centering
\caption{Quantitative results for the 10 simulation scenes from the KAIST dataset. PSNR (dB), SSIM and SAM are reported.}
\label{tab:Result_KAIST}
\renewcommand\arraystretch{1.05}
\setlength{\tabcolsep}{4.5pt}
\begin{tabular}{lc|ccccccccccc|lll}
\toprule
Method & Metrics  & S1    & S2    & S3    & S4    & S5    & S6    & S7    & S8    & S9    & S10   & Avg & \#Params & FLOPs  & \RD{Runtime} \\ \midrule

\multirow{2}{*}{GAP-TV~\cite{yuan2016generalized}}&PSNR & 26.82 & 22.89 & 26.31 & 30.65 & 23.64 & 21.85 & 23.76 & 21.98 & 22.63 & 23.10 & 24.36 &\multirow{2}{*}{-}&\multirow{2}{*}{-} &\multirow{2}{*}{-}\\ 
&SSIM & 0.754 & 0.610 & 0.802 & 0.852 & 0.703 & 0.663 & 0.688 & 0.655 & 0.682 & 0.584 & 0.669 && \\ \hline

\multirow{2}{*}{DeSCI~\cite{liu2018rank}} & PSNR & 27.13 & 23.04 & 26.62 & 34.96 & 23.94 & 22.38 & 24.45 & 22.03 & 24.56 & 23.59 & 25.27&\multirow{2}{*}{-}&\multirow{2}{*}{-} &\multirow{2}{*}{-} \\
& SSIM & 0.748 & 0.620 & 0.818 & 0.897 & 0.706 & 0.683 & 0.743 & 0.673 & 0.732 & 0.587 & 0.721 &&\\ \hline

\multirow{3}{*}{$\lambda$-Net~\cite{miao2019net}} & PSNR & 30.10 & 28.49 & 27.73 & 37.01 & 26.19 & 28.64 & 26.47 & 26.09 & 27.50 & 27.13 & 28.53&\multirow{3}{*}{62.64M}&\multirow{3}{*}{117.98G} &\multirow{3}{*}{\RD{42.41ms}} \\
& SSIM & 0.849 & 0.805 & 0.870 & 0.934 & 0.817 & 0.853 & 0.806 & 0.831 & 0.826 & 0.816 & 0.841&& \\
 & SAM &14.12&17.40&15.60&24.03&16.35&26.03&14.07&27.57&15.86&26.02&19.71&& \\  
\hline

\multirow{3}{*}{TSA-Net~\cite{meng2020end}}                 & PSNR & 32.03 & 31.00 & 32.25 & 39.19 & 29.39 & 31.44 & 30.32 & 29.35 & 30.01 & 29.59 & 31.46&\multirow{3}{*}{44.25M}&\multirow{3}{*}{110.06G} &\multirow{3}{*}{\RD{38.32ms}} \\
 &SSIM & 0.892 & 0.858 & 0.915 & 0.953 & 0.884 & 0.908 & 0.878 & 0.888 & 0.890 & 0.874 & 0.894& & \\
  & SAM & 8.73&10.35&7.39	&8.36&6.72&9.70&7.65	&11.37&7.66&9.55&8.75&&\\
 \hline  

\multirow{3}{*}{GAP-Net~\cite{meng2020gap}} & PSNR & 33.74 &33.26&  34.28  &41.03 & 31.44 & 32.40 & 32.27  &30.46 & 33.51 & 30.24 &33.26 &\multirow{3}{*}{4.27M}&\multirow{3}{*}{78.58G} &\multirow{3}{*}{\RD{49.87ms}}
 \\
& SSIM & 0.911 &0.900 &0.929 &0.967& 0.919 &0.925&  0.902  &0.905 & 0.915 & 0.895  &0.917&&\\
& SAM &9.11&13.07&8.60&9.54&7.87&12.61&	8.40&16.08&	8.75&	13.23	&10.73&&\\ 
 \hline 

\multirow{3}{*}{MST-L~\cite{cai2022mask}} & PSNR & 35.40 & 35.87 & 36.51 & 42.27 & 32.77 & 34.80 & 33.66 & 32.67 & 35.39 & 32.50 & 35.18 &\multirow{3}{*}{1.33M}&\multirow{3}{*}{19.42G} &\multirow{3}{*}{\RD{45.68ms}} \\
& SSIM & 0.941 & 0.944 & 0.953 & 0.973 & 0.947 & 0.955 & 0.925 & 0.948 & 0.949 & 0.941 & 0.948 && \\ 
 & SAM & 7.02&8.12&	6.08&7.42&	5.80&	7.84&6.27&	10.33&	7.46&8.35&	7.47&& \\  
\hline

%%%-------------------------
\multirow{3}{*}{MST-Plus~\cite{cai2022mask}} & PSNR & \RD{35.80}  & 36.23 & 37.34 & 42.63 & 33.38 & 35.38 & 34.35 & 33.71 & 36.67 & 33.38 & 35.99 
&\multirow{3}{*}{2.03M}&\multirow{3}{*}{28.15G} &\multirow{3}{*}{\RD{80.85ms}} \\
& SSIM & 0.943 & 0.947 & 0.957 & 0.973 & 0.952 & 0.957 & 0.934 & 0.953 & 0.953 & 0.945 & 0.951 && \\ 
& SAM & 6.74  & 7.47  & 4.59  & 6.95  & 4.64  & 6.61  & 5.92  & 8.29  & 5.72  & 6.62  & 6.35 && \\  
\hline

\multirow{3}{*}{CST-L~\cite{cai2022coarse}} & PSNR & 35.82 & 36.54 & 37.39 & 42.28 & 33.40 & 35.52 & 34.44 & 33.83 & 35.92 & 33.36 & 35.85 &\multirow{3}{*}{3.00M}&\multirow{3}{*}{27.81G} &\multirow{3}{*}{\RD{70.33ms}} \\
& SSIM & 0.947 & 0.952 & 0.959 & 0.972 & 0.953 & 0.962 & 0.937 & 0.959 & 0.951 & 0.948 & 0.954 &&\\
 & SAM & 6.60	&6.76&	4.71&	5.78&	4.45&	5.95&	5.64&	6.82&	5.70&	5.89&	5.83 &&\\ 
\hline

%%%-------------------------
\multirow{3}{*}{CST-Plus~\cite{cai2022coarse}} & PSNR & 35.96 & 36.84 & 38.16 & 42.44 & 33.25 & 35.72 & 34.86 & 34.34 & 36.51 & 33.09 & 36.12 &\multirow{3}{*}{3.00M}&\multirow{3}{*}{40.10G} &\multirow{3}{*}{\RD{90.56ms}} \\
& SSIM & 0.949 & 0.955 & 0.962 & 0.975 & 0.955 & 0.963 & 0.944 & 0.961 & 0.957 & 0.945 & 0.957 &&\\
& SAM & \RD{6.40}   & 6.46  & 4.26  & 5.78  & 4.69  & 5.91  & 5.43  & \RD{6.70}   & 5.46  & 5.96  & 5.71 &&\\ 
\hline

%%%-------------------------
\multirow{3}{*}{DAUHST~\cite{cai2022degradation}} & PSNR & 36.59 & 37.93 & 39.32 & \textbf{44.77} & 34.82 & 36.19 & 36.02 & 34.28 & \textbf{38.54} & 33.67 & 37.21
  &\multirow{3}{*}{2.08M} &\multirow{3}{*}{27.17G} &\multirow{3}{*}{\RD{63.05ms}} \\
& SSIM & 0.949 & 0.958 & 0.964 & 0.980  & 0.961 & 0.963 & 0.95  & 0.956 & 0.963 & 0.947 & 0.959 &&\\ 
& SAM & 5.79 & 6.75  & 3.85  & 5.36  & 3.95  & 6.04  & 4.86  & 7.97  & 4.78  & 6.04  & 5.54 &&\\ 
\hline

\multirow{3}{*}{PADUT~\cite{li2023pixel}} & PSNR & 36.25 & 37.92 & \textbf{39.63} & 44.55 & 34.59 & 35.58 & 35.69 & 33.76 & 38.26 & 33.24   & 36.95 &\multirow{3}{*}{1.35M} &\multirow{3}{*}{22.91G} &\multirow{3}{*}{\RD{81.86ms}} \\
& SSIM &  0.951 & 0.963 & \textbf{0.970} & \textbf{0.985} & 0.964 & 0.965 & 0.950 & 0.960 & 0.963 & 0.947 &  0.962 &&\\ 
 & SAM &  5.79&	5.69&3.55&	4.03 & \textbf{3.48} &	5.04&4.91&	6.59&	\textbf{4.26}	&5.09&4.84 &&\\ 
\midrule

\multirow{3}{*}{\textbf{LSST-S (Ours)}} & PSNR & 35.27 & 35.49 & 36.95 & 42.64 & 32.45 & 34.07 & 33.72 & 32.70  & 36.23 & 32.32 & 35.18 &\multirow{3}{*}{0.69M} &\multirow{3}{*}{8.37G} &\multirow{3}{*}{\RD{18.83ms}} \\
& SSIM & 0.944 & 0.939 & 0.961 & 0.98  & 0.943 & 0.952 & 0.935 & 0.948 & 0.957 & 0.939 & 0.949 &&\\ 
& SAM & 6.89  & 7.83  & 5.01  & 7.04  & 5.37  & 8.01  & 5.98  & 8.34  & 6.44  & 7.08  & \RD{6.80} &&\\ 
\hline

\multirow{3}{*}{\textbf{LSST-M (Ours)}} & PSNR & 35.59 & 36.63 & 37.50  & 41.85 & 33.18 & 35.09 & 34.00    & 33.24 & 36.97 & 32.81 & 35.68 &\multirow{3}{*}{0.85M} &\multirow{3}{*}{13.04G} &\multirow{3}{*}{\RD{27.14ms}} \\
& SSIM & 0.947 & 0.954 & 0.964 & 0.975 & 0.954 & 0.962 & 0.941 & 0.959 & 0.963 & 0.948 & 0.956 &&\\ 
& SAM & 6.86  & 6.27  & 4.79  & 6.56  & \RD{4.50}   & 6.44  & 5.59  & 7.27  & 5.25  & \RD{6.60}   & 6.01 &&\\ 
\hline

\multirow{3}{*}{\textbf{LSST-L (Ours)}} & PSNR & 35.93 & 37.32 & 38.60  & 44.06 & 33.66 & 35.87 & 35.07 & 33.70  & 37.35 & 33.54 & 36.51 &\multirow{3}{*}{1.22M} &\multirow{3}{*}{16.35G} &\multirow{3}{*}{\RD{34.49ms}} \\
& SSIM & 0.953 & 0.962 & \textbf{0.970}  & 0.984 & 0.961 & 0.968 & 0.948 & 0.966 & 0.966 & 0.957 & 0.963 &&\\ 
& SAM & \textbf{5.69}  & 6.13  & 3.88  & 4.87  & 3.97  & 5.31  & 5.06  & 6.43  & 4.96  & 5.30   & 5.16 &&\\ 
\hline

\multirow{3}{*}{\textbf{LSST-Plus (Ours)}} & PSNR & \textbf{37.13} & \textbf{38.48} & 39.17 & 44.36 & \textbf{35.03} & \textbf{36.85} & \textbf{36.38} & \textbf{34.76} & 37.69 & \textbf{34.48} & \textbf{37.43} &\multirow{3}{*}{1.35M} &\multirow{3}{*}{22.60G} &\multirow{3}{*}{\RD{40.43ms}} \\
& SSIM & \textbf{0.958} & \textbf{0.965} & 0.965 & \textbf{0.985} & \textbf{0.965} & \textbf{0.972} & \textbf{0.956} & \textbf{0.969} & \textbf{0.964} & \textbf{0.960} & \textbf{0.966} &&\\ 
& SAM & 5.88  & \textbf{5.58} & \textbf{3.40} & \textbf{3.83} & 3.71  & \textbf{4.62} & \textbf{4.84} & \textbf{5.50} & 4.47  & \textbf{4.59} & \textbf{4.64} &&\\ 

\bottomrule
\end{tabular}%
\end{table*}

\section{Experiment}

\subsection{Data and Experimental Setups}
We offer a thorough evaluation of the proposed LSST method across multiple datasets, including those specifically designed for indoor scenes such as CAVE~\cite{park2007multispectral} and KAIST~\cite{choi2017high}, as well as datasets covering both indoor and outdoor scenarios like ICVL~\cite{arad2016sparse} and Harvard~\cite{chakrabarti2011statistics}. We followed the experimental protocols set by TSA-Net~\cite{meng2020end} to ensure consistent comparisons across both simulated and real datasets. In the simulations, we assessed the model on 10 selected scenes from KAIST, using CAVE as the training set. To test the algorithm’s generalization capabilities, we conducted evaluations on the Harvard and ICVL datasets, randomly selecting 10 scenes for testing and using the remaining scenes for training.

In real dataset experiments, we used actual hyperspectral images captured by the CASSI system as outlined by TSA-Net~\cite{meng2020end}. Training commenced from scratch on the CAVE and KAIST datasets, incorporating 11-bit grain noise to emulate two-dimensional compressed mixed images during training. Subsequently, we performed tests on real datasets, which included five different scenes of two-dimensional compressed mixed images, each with dimensions $660 \times 714$.

In this study, all models were trained over 300 epochs using the Adam optimizer, with an initial learning rate of $4 \times 10^{-4}$.  For experiments involving simulated data, we extracted spatial patches of $256 \times 256$ from the simulated hyperspectral images for model input. In the case of real hyperspectral image reconstruction, the patch size was increased to $660 \times 660$ to accommodate the actual dimensions of two-dimensional mixed images. The disperser's step size was set at 2, yielding measurement sizes of $256 \times 310$ for simulated data and $660 \times 714$ for real data. We maintained a batch size of 5, employing data augmentation methods such as random flipping and rotation. The models LSST-S, LSST-M, and LSST-L were configured with LSSTB repeat counts of 1, 1, 2; 2, 2, 2; and 2, 3, 3, respectively. Additionally, the LSST-Plus model combined three LSST-S models into a cohesive three-step end-to-end network. \RD{The number of groups ${\rm G}$ was set to 4.} Both training and testing of these models were performed on an RTX 3090 GPU.

We employed four quantitative image quality metrics to assess the performance of all methods: Peak Signal-to-Noise Ratio (PSNR) and Structural Similarity Index (SSIM). Higher values of PSNR and SSIM signify improved performance.

\subsection{Results on CAVE and KAIST Datasets}
We compared our LSST algorithm against several leading reconstruction algorithms. \RD{These included model-based method—DeSCI~\cite{liu2018rank};} three CNN-based methods—$\lambda$-Net~\cite{miao2019net}, TSA-Net~\cite{meng2020end}, and GAP-Net~\cite{meng2020gap}; and four Transformer-based methods—MST~\cite{cai2022mask}, CST~\cite{cai2022coarse}, DAUHST~\cite{cai2022degradation}, and PADUT~\cite{li2023pixel}. To ensure a fair comparison, all methods were evaluated under the same conditions as established for TSA-Net.

\begin{figure*}[t]
    \centering{\includegraphics[width=0.96\textwidth]{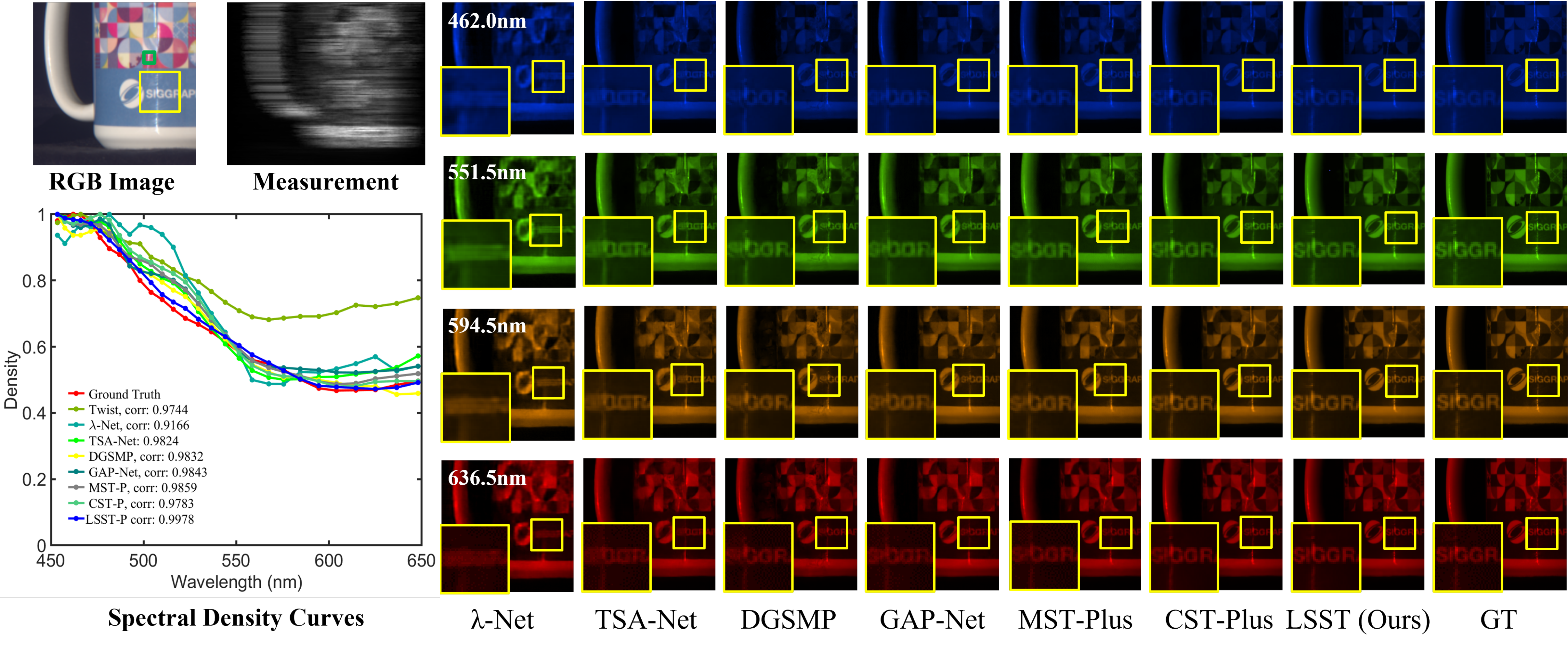}}
    \vspace{-3mm}
    \caption{Visualization of reconstruction results for Scene 5 from the KAIST Dataset. The reconstructed spectra curves for different methods in the region highlighted by the \textcolor{green}{green} box within the RGB image are also presented.}
	\label{fig:Vis_KAIST}
\end{figure*}

\begin{figure}[t]
    \centering
    \includegraphics[width=0.5\textwidth]{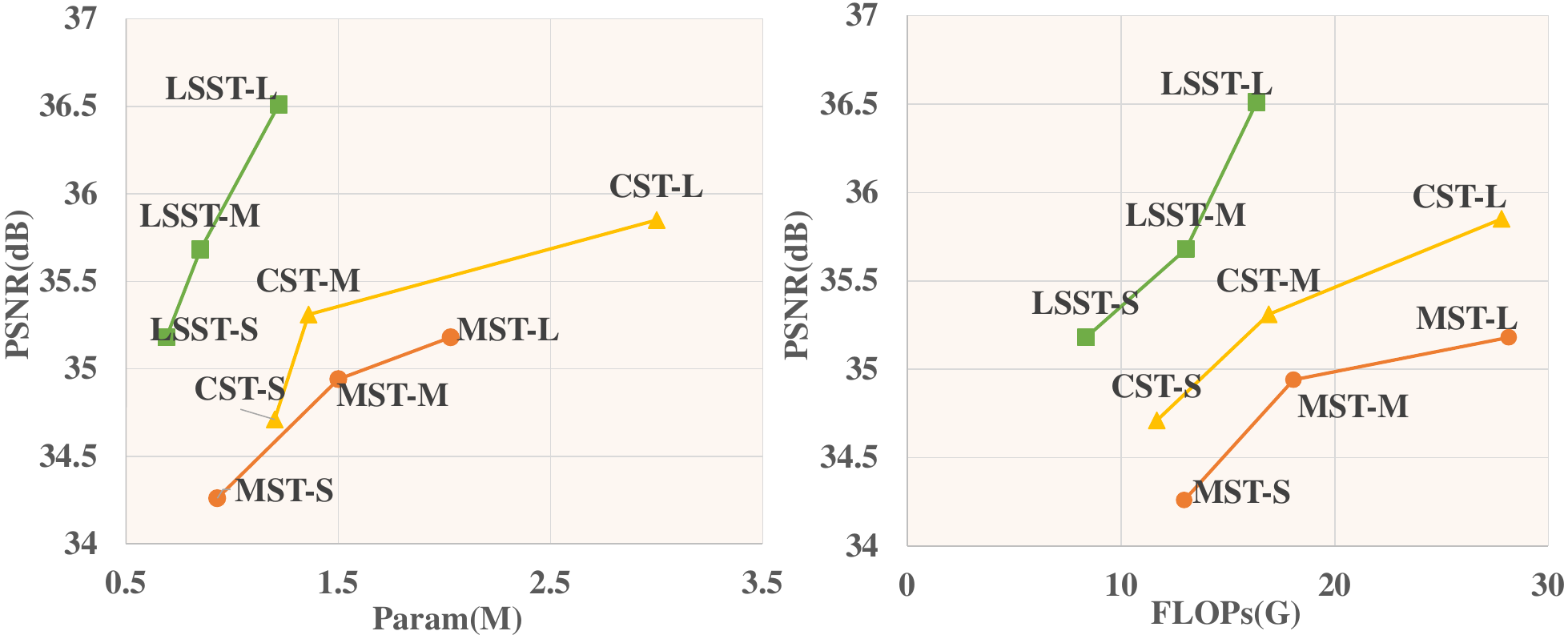}
    \vspace{-6mm}
    \caption{Comparison of PSNR-Param and PSNR-FLOPs for LSST, MST, and CST.}
    \vspace{-3mm}
    \label{fig:Vis_Flops}
\end{figure}

\noindent\textbf{Quantitative Result}. 
The performance of various methods is detailed in~\cref{tab:Result_KAIST}. Significantly, LSST-Plus offers marked improvements over previous methods in reconstruction quality across all 10 scenarios. Additionally, it achieves a reduction in the number of parameters (Params) and computational costs (FLOPs). Specifically, compared to established CNN-based methods such as TSA-Net~\cite{meng2020end}, LSST-Plus shows an enhancement of $5.97$ dB in PSNR and $0.072$ in SSIM. \RD{This improvement accompanies a significant reduction in resource utilization}, with only $3.1\%$ of the parameters (1.35 M out of 44.25 M) and $20.5\%$ of the FLOPs (22.60 G out of 110.06 G) required.

Moreover, compared to Transformer-based methods, MST-Plus~\cite{cai2022mask} and CST-Plus~\cite{cai2022coarse}, LSST-Plus demonstrates enhancements of $1.44$ dB and $1.31$ dB in PSNR, respectively. These improvements are achieved with significantly lower resource utilization, with only $66.5\%$ (\RD{1.35 M out of 2.03 M}) and $45.0\%$ (1.35 M out of 3.00 M) of parameters, and $80.3\%$ (22.60 G out of 28.15 G) and $56.4\%$ (22.60 G out of 40.10 G) of FLOPs, respectively. Furthermore, when compared to the recent Transformer-based method PADUT~\cite{cai2022degradation}, LSST-Plus achieves a $0.48$ dB increase in PSNR while maintaining equivalent parameter counts and utilizing fewer FLOPs.

\begin{table}[t]
    \vspace{-1mm}
    \caption{Quantitative results on Harvard and ICVL dataset.}
    \label{tab:Result_Harvard}
    \centering
    \renewcommand\arraystretch{1.1}
    \setlength{\tabcolsep}{6.0pt}
    \begin{tabular}{l|ccc|ccc}
    \toprule
    \multirow{2}{*}{Method} & \multicolumn{3}{c|}{\textbf{Harvard}} & \multicolumn{3}{c}{\textbf{ICVL}} \\ \cline{2-7}
    & PSNR & SSIM & SAM & PSNR & SSIM & SAM \\
    \midrule
    $\lambda$-Net~\cite{miao2019net} & 34.78 & 0.877 & 5.04 & 37.28 & 0.934 & 4.64 \\
    TSA-Net~\cite{meng2020end} & 35.62 & 0.884 & 4.81 & 39.14 & 0.949 & 3.29 \\
    GAP-Net~\cite{meng2020gap} & 36.04 & 0.890  & 4.30 & 39.27 & 0.949 & 3.28 \\
    MST-L~\cite{cai2022mask} & 37.57 & 0.918 & 3.64 & 41.23 & 0.956 & 2.62 \\
    CST-L~\cite{cai2022coarse} & 37.73 & 0.921 & 3.43 & 41.58 & 0.957 & 2.44 \\ 
    \midrule
    \rowcolor{RowColor} \textbf{LSST-L (Ours)} & \textbf{38.24} & \textbf{0.927} & \textbf{3.20} & \textbf{42.39} & \textbf{0.958} & \textbf{2.14} \\
    \bottomrule
    \end{tabular}%
   \label{tab:3.6}%
   \vspace{-1mm}
\end{table}%

\noindent\textbf{Efficiency Analysis}. 
Additionally, we compare the Params and FLOPs of our LSST models—small, medium, large and plus—with end-to-end Transformer models MST and CST as illustrated in \cref{fig:Vis_Flops}. These comparisons reveal that our proposed LSST models achieve superior reconstruction performance while requiring fewer parameters and computational resources.

\RD{Furthermore, we evaluated the inference time of a broad set of deep learning algorithms, and the statistical results are summarized in \cref{tab:Result_KAIST}. Our LSST models consistently outperform other Transformer-based methods in runtime. Although PADUT~\cite{li2023pixel} has comparable computational complexity to LSST, its deep-unfolding optimization framework leads to noticeably longer inference times. In contrast, LSST benefits from its ultra-low computational complexity and streamlined end-to-end architecture, resulting in significantly shorter inference latency. This advantage is particularly valuable for practical hyperspectral compression imaging, where efficient reconstruction is critical.}

\noindent\textbf{Visualization Result}.
\cref{fig:Vis_KAIST} presents reconstructions of simulated hyperspectral images from Scene 5 using seven state-of-the-art methods alongside our proposed LSST-Plus method, focusing on 4 out of 28 spectral channels. The analysis of the reconstructed images and the enlarged areas within the yellow boxes highlights the limitations of previous methods in restoring such images. Previous techniques often result in overly smooth images that lack fine structural and textural details, or they introduce unwanted color artifacts and patchy textures. In contrast, LSST-Plus exhibits superior perceptual image quality, providing clearer textural details, enhanced noise suppression, and greater fidelity.

Additionally, \cref{fig:Vis_KAIST} includes plots of spectral curves corresponding to the yellow boxed area. LSST-Plus demonstrates the most accurate spectral curve reconstruction, exhibiting the highest correlation with the authentic curves, thus confirming the effectiveness of LSST-Plus.

\begin{figure}[t]
    \centering
    \includegraphics[width=0.49\textwidth]{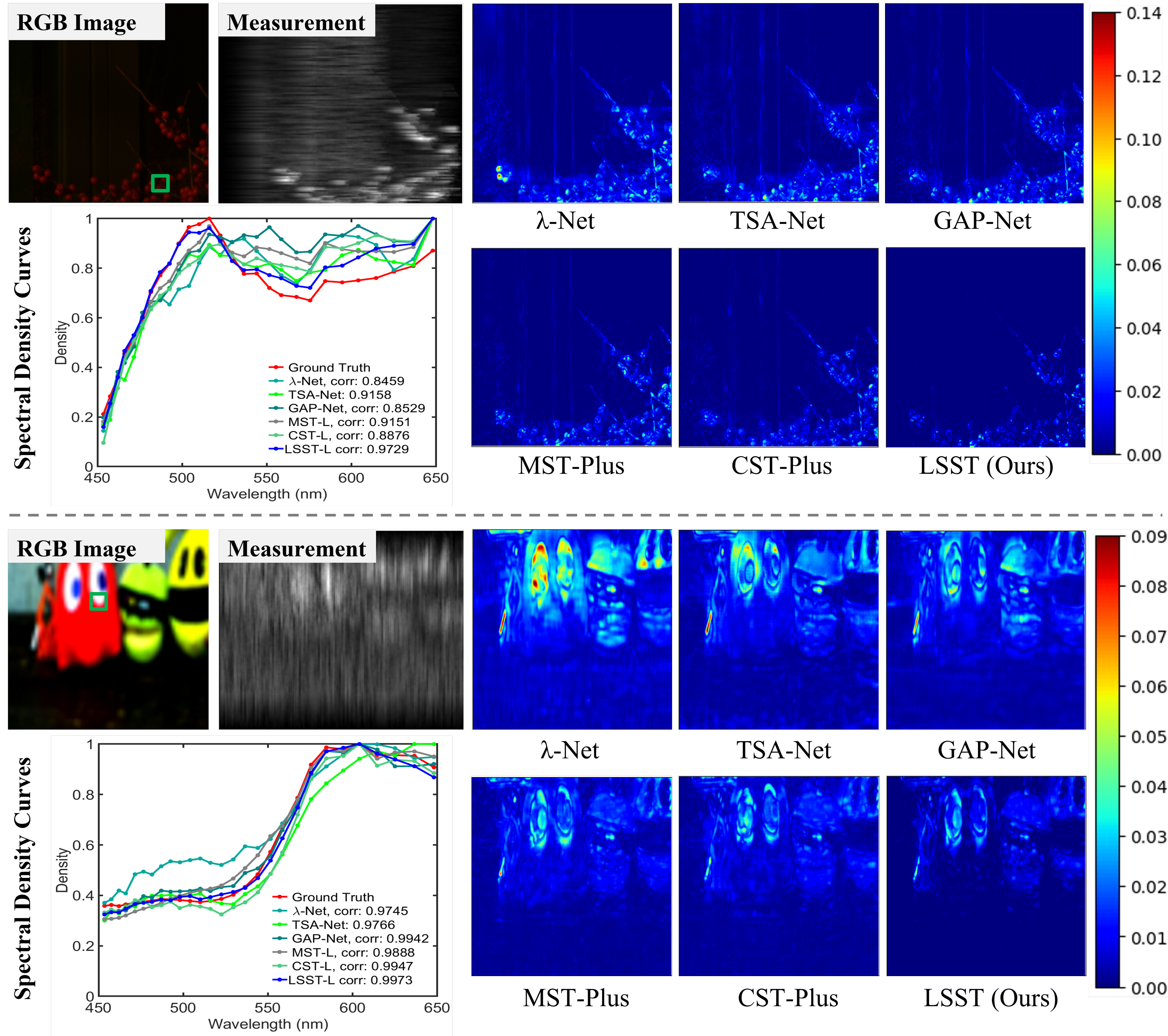}
    \vspace{-5mm}
    \caption{Visualization of reconstruction error maps and the reconstructed spectra curves for different methods in the region highlighted by the \textcolor{green}{green} box within the RGB image of the Harvard (top) and the ICVL (bottom) dataset.}
    \label{fig:Vis_Error}
\end{figure}

\subsection{Results on Harvard and ICVL Datasets}

\noindent\textbf{Quantitative Result.}
To substantiate the generalization capabilities of the proposed model, this study evaluated its performance on the Harvard and ICVL datasets, which include outdoor scenes. We compared the performance of LSST-L with five advanced methods. Quantitative results, as shown in \cref{tab:Result_Harvard}, reveal that LSST-L outperforms the competing methods. Specifically, on the Harvard dataset, LSST-L achieved higher PSNR than $\lambda$-Net~\cite{miao2019net}, TSA-Net~\cite{meng2020end}, GAP-Net~\cite{meng2020gap}, MST-L~\cite{cai2022mask}, and CST-L~\cite{cai2022coarse} by margins of $3.46$, $2.62$, $2.20$, $0.67$, and $0.51$, respectively. Similarly, on the ICVL dataset, LSST-L demonstrated improvements in PSNR of $5.11$, $3.25$, $3.12$, $1.16$, and $0.81$, respectively. These results affirm that the LSST algorithm surpasses state-of-the-art methods, thereby confirming its robustness and generalization capabilities for outdoor scenes.

\noindent\textbf{Visualization Result.}
\cref{fig:Vis_Error} depicts the reconstruction errors in scenes from the Harvard and ICVL test sets, where different colors represent varying error levels and blue signifies lower errors. It is clear that our method consistently achieves smaller reconstruction errors. This is particularly noticeable in areas with rich spectral features, such as the eye region of the left object in the ICVL dataset, where our approach significantly reduces errors.

Furthermore, \cref{fig:Vis_Error} also presents a comparison of spectral reconstruction curves across different methods for both Harvard and ICVL test scenes. On the Harvard dataset, our method markedly excels, achieving a correlation coefficient of $0.973$, compared to a maximum of $0.916$ achieved by other methods. Similarly, on the ICVL dataset, our algorithm exhibits the closest correlation to the true spectral curve, achieving a coefficient of $0.997$.

\begin{table}[t]
   \caption{Ablation study on key network components. FSL: Focal Spectrum Loss.}
   \label{tab:ab_module}
    \centering
    \renewcommand\arraystretch{1.2}
    \setlength{\tabcolsep}{7.2pt}
    \begin{tabular}{ccc|cccc}
    \toprule
    LSCB  & SS-MSA & FSL   & PSNR & SSIM & \#Param & FLOPs \\
    \midrule
    \xmark  &  \xmark  &  \xmark   &   32.04 & 0.894 & 0.32M  & 4.33G  \\
    \cmark  &  \xmark  &  \xmark   &   33.35 & 0.937 & 0.67M  & 7.75G \\
    \cmark  &  \cmark  &  \xmark   &   34.83 & 0.944 & 0.69M  & 8.37G \\
    \rowcolor{RowColor} \cmark  &  \cmark  &  \cmark   &   \textbf{35.18} & \textbf{0.949} & 0.69M  & 8.37G \\
    \bottomrule
    \end{tabular}
    \label{tab:3.8}
\end{table}

\begin{table}[t]
\caption{Ablation Study on Multi-Head Self-Attention.}
\label{tab:abl_MSA}
\centering
\renewcommand\arraystretch{1.2}
\setlength{\tabcolsep}{1.0pt}
\begin{tabular}{c|cccccc >{\columncolor{RowColor}}c}
\toprule
Method & Baseline & G-MSA  &  W-MSA  & Swin-MSA  & S-MSA  & SAH-MSA  & SS-MSA  \\
\midrule
PSNR  & 32.04 & 33.42 & 33.53 & 33.62 & 33.86 & 34.11 & \textbf{34.67} \\
SSIM  & 0.894 & 0.932 & 0.936 & 0.937 & 0.939 & 0.942 & \textbf{0.945} \\
\#Params & 0.32M  & 0.45M  & 0.45M  & 0.45M  & 0.45M  & 0.45M  & 0.34M \\
FLOPs & 4.33G  & 4.58G  & 5.66G  & 5.66G  & 5.62G  & 5.54G  & 5.12G \\
\bottomrule
\end{tabular}
\label{tab:3.9}
\end{table}

\subsection{Results on Real Dataset}
\begin{figure*}[t]
    \centering{\includegraphics[width=0.85\textwidth]{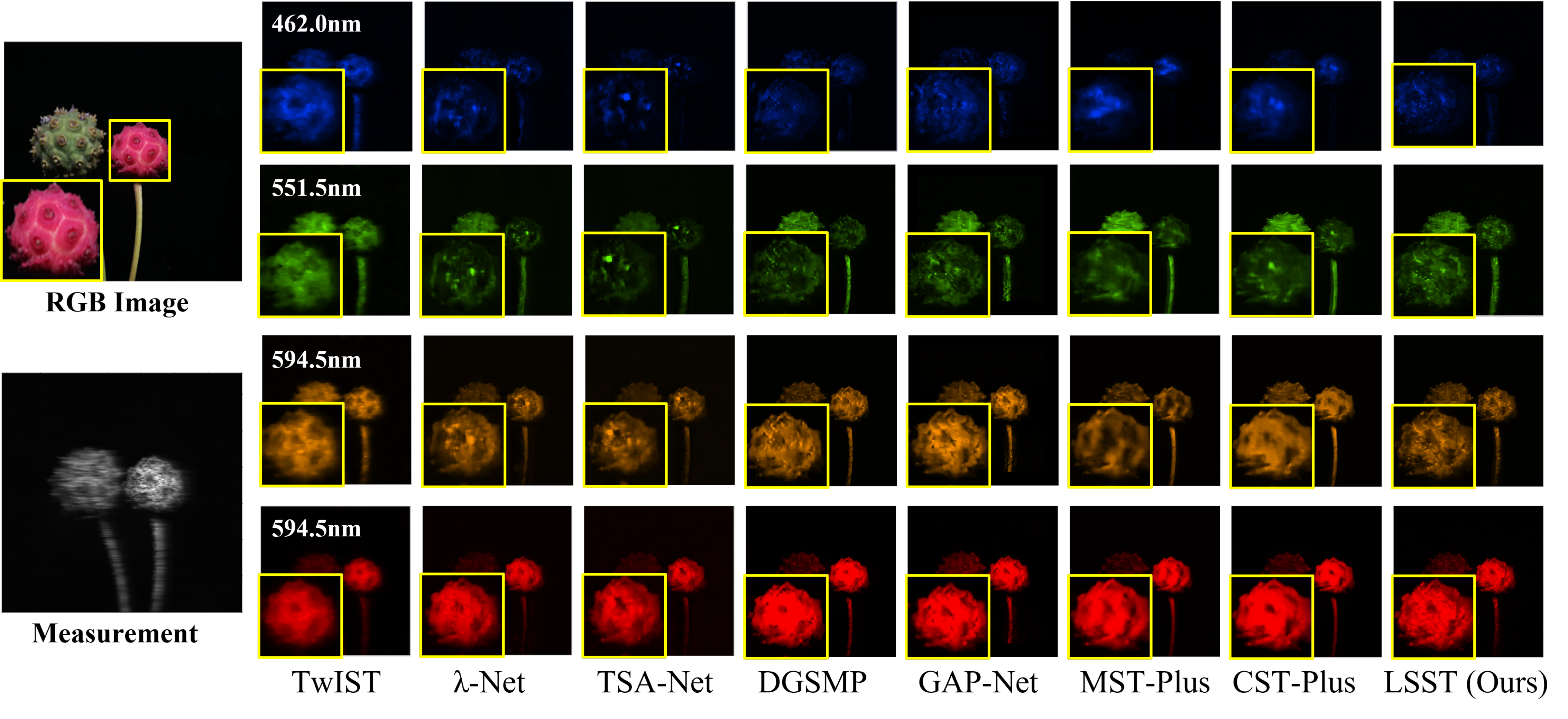}}
    \vspace{-3mm}
    \caption{Visualization of reconstruction results from various methods applied to real-world scenes.}
    \label{fig:Vis_Real}
    \vspace{-2mm}
\end{figure*}

Due to the absence of ground-truth data in the real hyperspectral image dataset, quantitative evaluation metrics could not be applied. Consequently, we focused on comparing the qualitative results of our LSST approach with other methods. \cref{fig:Vis_Real} displays the reconstruction results from one randomly selected scene out of five real scenes. Previous methods managed only to reconstruct rough outlines, often resulting in excessive smoothing and distortion of details. In contrast, LSST-Plus recovers more textures and details, yielding visually more appealing results. The reconstruction outcomes of LSST-Plus demonstrate enhanced noise suppression and maintain higher visual fidelity. These results on the real dataset underscore the robustness and generalization capability of the LSST-Plus model.

\subsection{Ablation Study}
\noindent\textbf{Effect of Key Modules.}
We conducted ablation studies to assess the individual contributions of various components in our model, as depicted in~\cref{tab:3.8}. The baseline model, derived by omitting the Lightweight Spatial Convolution Block (LSCB), Separate Spectral Multi-head Self-Attention (SS-MSA), and Focal Spectrum Loss (FSL) from LSST-S, achieved a PSNR of $32.04$ dB and an SSIM of $0.894$. The inclusion of LSCB improved the PSNR by $1.31$ dB and SSIM by $0.043$. Adding SS-MSA further increased the PSNR by $1.48$ dB and SSIM by $0.007$, with a marginal increase in parameters by $0.02$ M and computational complexity by $0.62$ G FLOPs, highlighting the efficiency of SS-MSA. Incorporation of FSL further enhanced the PSNR by $0.35$ dB and SSIM by $0.005$. These results collectively demonstrate the significant impact and effectiveness of the key components integrated within LSST.

\begin{figure}[t]
    \centering
    \includegraphics[width=0.465\textwidth]{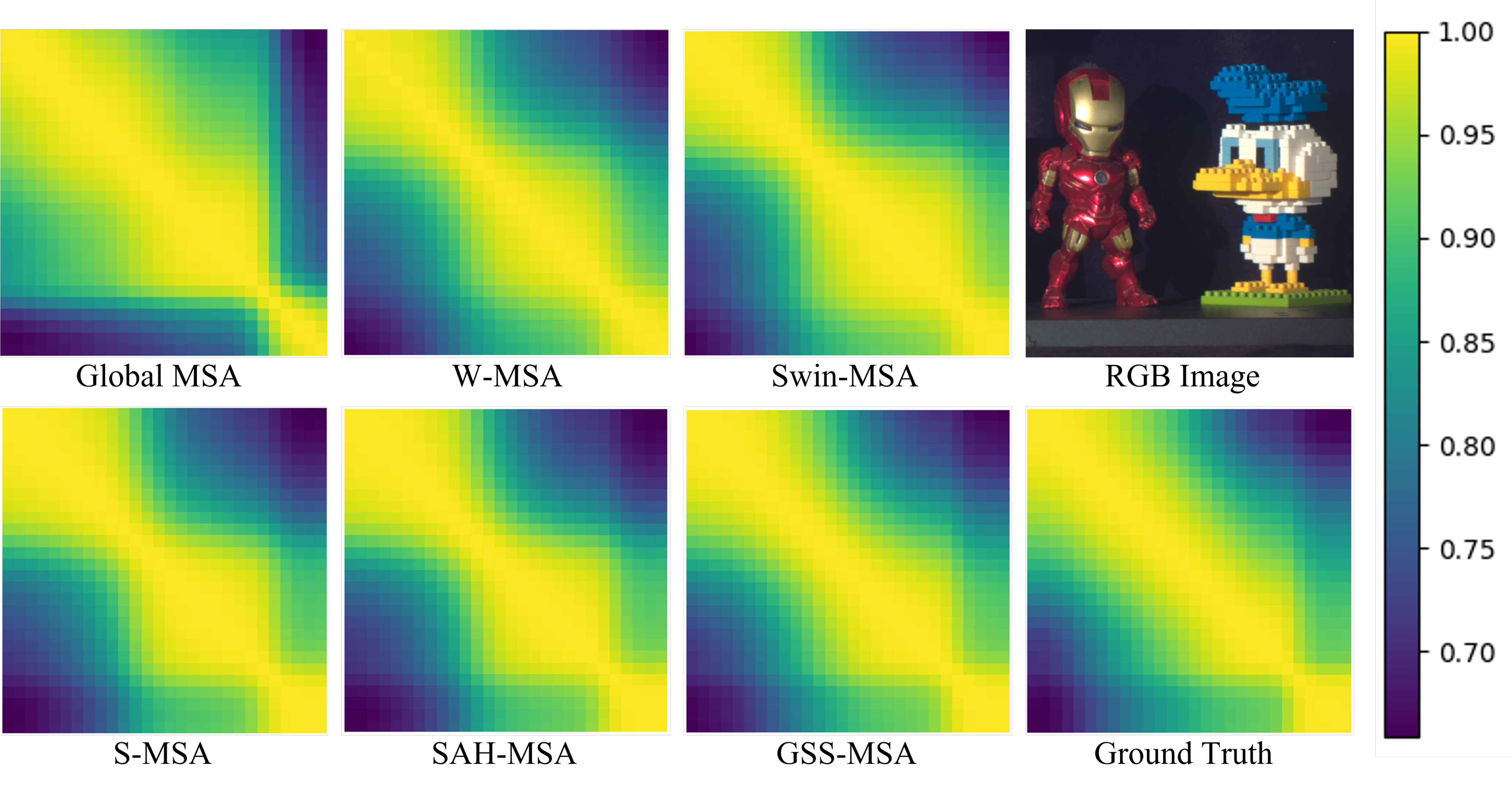}
    \vspace{-3mm}
    \caption{Visualization of correlation coefficient maps from various methods.}
    \label{fig:Vis_Corr}
\end{figure}

\noindent\textbf{Effect of SS-MSA.}
\RD{We further examine the effectiveness of our Separate Spectral Multi-head Self-Attention (SS-MSA) design. \cref{tab:abl_MSA} first reports the performance of the baseline model (configured as described in the previous paragraph), which achieves a PSNR of $32.04$ dB and an SSIM of $0.894$, with $0.32$ M parameters and $4.33$ G FLOPs. Based on this baseline, we evaluated various attention mechanisms, including Global Multi-Head Self-Attention (G-MSA), Window Multi-Head Self-Attention (W-MSA), Swin Multi-Head Self-Attention (Swin-MSA), Spectral Attention (S-MSA), Spectral-aware Hash Clustering Multi-Head Self-Attention (SAH-MSA), and our proposed SS-MSA. Notably, G-MSA requires halving the input features to alleviate memory constraints.}

\RD{Among all variants, SS-MSA achieves the most significant performance improvements with only a marginal increase in parameters and computational cost. Specifically, it adds only $0.02$ M parameters and $0.79$ G FLOPs, yet improves PSNR by $2.63$ dB and SSIM by $0.051$. These gains can be attributed to SS-MSA’s ability to efficiently model both local and non-local spectral dependencies in high-dimensional spectral data. In addition, its design ensures that computational complexity scales linearly with both spatial and spectral dimensions, minimizing redundant computations and highlighting the computational and memory efficiency of SS-MSA.}

\RD{Furthermore, the visual analysis in \cref{fig:Vis_Corr} shows the correlation coefficients among the spectra of hyperspectral images reconstructed by MSA models with different mechanisms. The analysis is performed by flattening each spectral band into a 1D vector and computing the pairwise correlation coefficients. Models equipped with SS-MSA produce correlation maps that closely resemble the ground truth, clearly demonstrating the ability of SS-MSA to capture complex spectral relationships.}

\begin{figure}[t]
    \centering
    \includegraphics[width=0.38\textwidth]{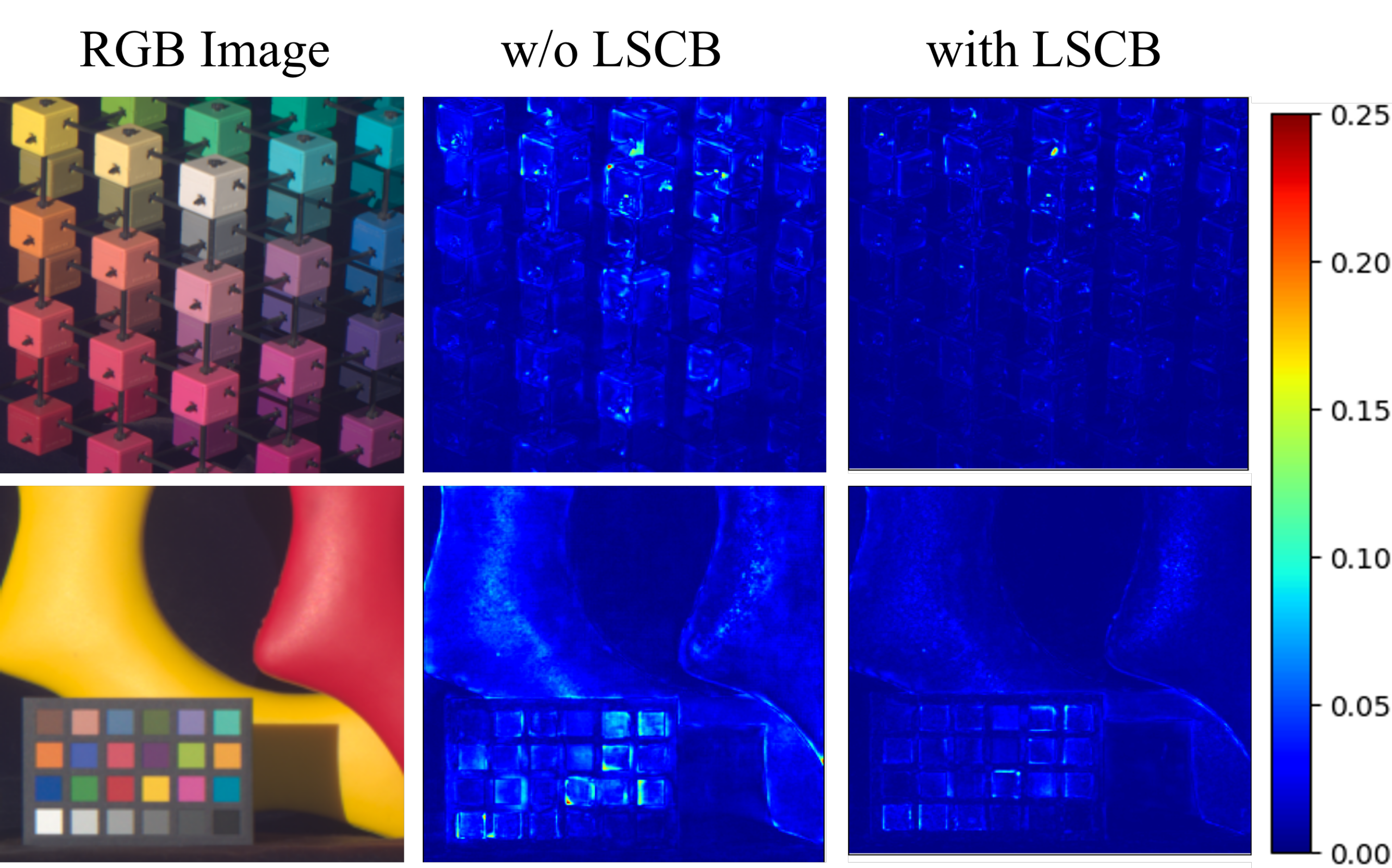}
    \vspace{-2mm}
    \caption{Comparison of reconstruction error maps with and without LSCB.}
    \vspace{-2mm}
    \label{fig:Vis_LSCB}
\end{figure}

\noindent\textbf{Effect of Lightweight Spatial Convolution.}
To more intuitively examine the benefits of the Lightweight Spatial Convolution Block (LSCB), we visualize error maps reconstructed by two models—one with and one without the LSCB module, as depicted in~\cref{fig:Vis_LSCB}. The use of the LSCB module clearly results in reduced reconstruction errors. Notably, in areas rich in spatial and spectral details, the integration of LSCB significantly improves regions that previously exhibited higher reconstruction errors. This module notably enhances the accurate reconstruction of key spatial structures, such as the upper region in the first image and the bottom-left corner of the second image. This enhancement primarily stems from LSCB's proficiency in local spatial modeling.

\begin{figure}[t]
    \centering
    \includegraphics[width=0.48\textwidth]{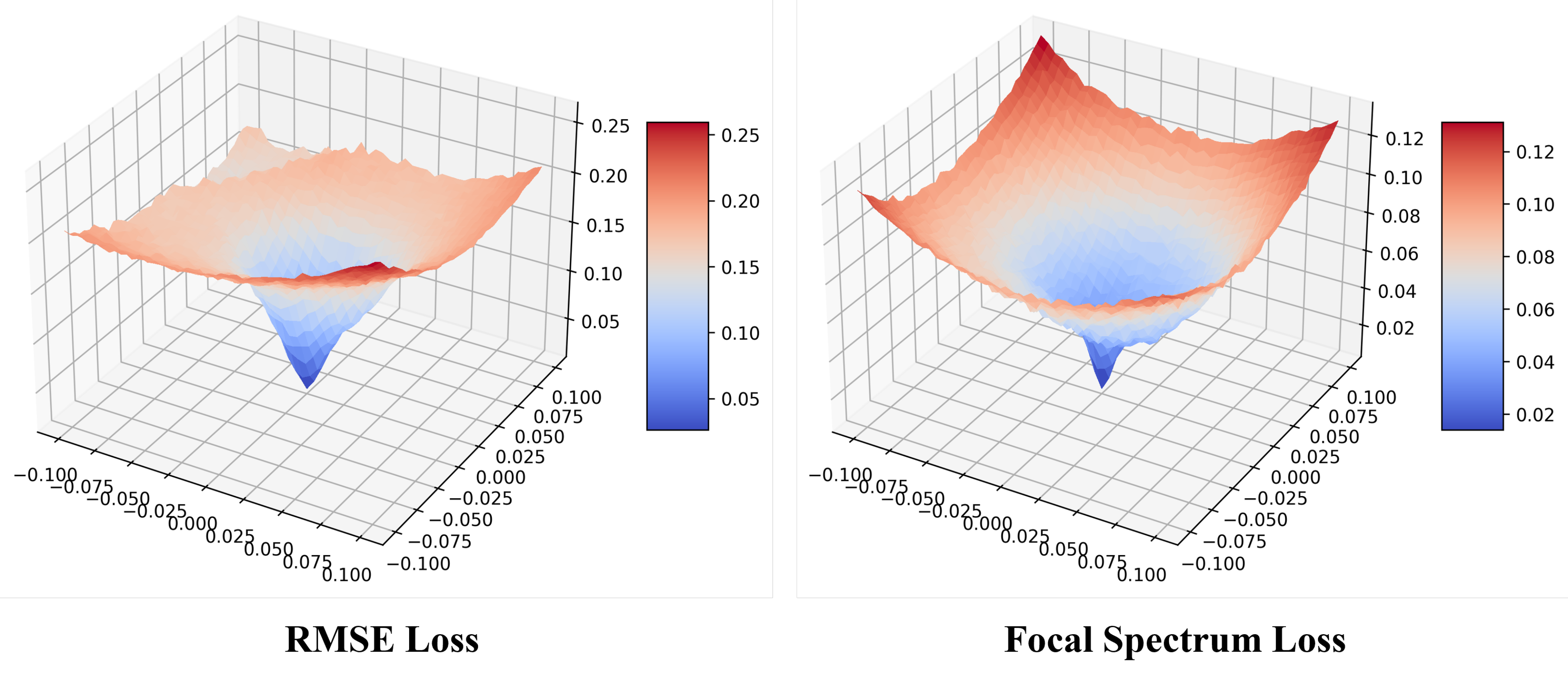}
    \vspace{-2mm}
    \caption{Comparison of loss landscape resulting from RMSE Loss and our Focal Spectrum Loss.}
    \label{fig:Landscape}
\end{figure}

\noindent\textbf{Effect of Focal Spectrum Loss.}
To further assess the effectiveness of Focal Spectrum Loss, \cref{fig:Landscape} displays the Root Mean Square Error (RMSE) loss alongside the loss landscape of the Focal Spectrum Loss. This landscape depicts the loss as a function of the neural network parameters $\alpha$. Notably, compared to the RMSE loss, the Focal Spectrum Loss surface exhibits a smoother profile, indicating fewer abrupt changes, while the RMSE surface appears more chaotic and irregular. These observations validate the efficacy of Focal Spectrum Loss in providing a more stable and consistent training process.

\section{Conclusion}
This study introduced the Lightweight Separate Spectral Transformer (LSST), a novel approach to hyperspectral image reconstruction. The LSST incorporates the Separate Spectral Transformer Block (SSTB) and the Lightweight Spatial Convolution Block (LSCB), effectively addressing the intricate spatial and spectral characteristics of hyperspectral data while substantially reducing computational demands. Additionally, the incorporation of Focal Spectrum Loss ensures balanced reconstruction quality across all spectral bands, enhancing accuracy. Experimental results validate LSST's superior performance and efficiency compared to existing methods, demonstrating its potential to enhance capabilities for resource-limited applications. 

\RD{Although our model achieves visually superior reconstruction quality on real-world data, noise in practical acquisition scenarios can be far more severe, complex, and non-stationary. Developing more robust noise modeling and reconstruction strategies therefore represents an important direction for future work. In addition, we plan to explore the feasibility of combining LSST with unfolding-based reconstruction paradigms to further investigate its potential in model-based reconstruction.}

\bibliographystyle{IEEEtran}
\bibliography{egbib}

\vfill

\end{document}